\documentclass[10pt,twocolumn]{article}

\PassOptionsToPackage{table,xcdraw}{xcolor}
\usepackage[preprint]{antintlpaper}
\usepackage[utf8]{inputenc}
\usepackage{CJKutf8}
\usepackage{inconsolata}
\usepackage{colortbl}
\usepackage{listings}
\usepackage{needspace}
\tcbuselibrary{listings,breakable,skins}
\DeclareCaptionFont{10pt}{\fontsize{10}{12}\selectfont}

\definecolor{HBTableGroup}{HTML}{F4F5F6}
\definecolor{HBTableMethod}{HTML}{F2F7FC}
\AntTitle{\texorpdfstring{Heartbeat Thinking: \\
Speculative Reasoning and Tool Execution 
While Listening in Realtime Voice Agents}{Heartbeat Thinking: Speculative Reasoning and Tool Execution While Listening in Realtime Voice Agents}}
\AntRunningTitle{Heartbeat Thinking}
\AntAuthors{\texorpdfstring{%
  \mbox{Zhenyu Zhang} \and
  \mbox{Kewei Fu} \and
  \mbox{Yongqi Tong} \and
  \mbox{Chenzhi Zheng} \and
  \mbox{Wei Wu} \and
  \mbox{Jiang-Ming Yang} \and
  \mbox{Jianshe Li}%
}{Zhenyu Zhang, Kewei Fu, Yongqi Tong, Chenzhi Zheng, Wei Wu, Jiang-Ming Yang, Jianshe Li}}
\AntAffiliations{Ant International}
\AntContact{\textbf{Correspondence:} tongyongqi.yq@ant-intl.com}
\AntDate{}
\AntKeywords{voice agents, speculative reasoning, tool execution}
\AntAbstract{Tool-using voice agents often leave users waiting after speech ends, even when actionable information arrives earlier in the request.
Speculative approaches begin work during speech, but chunk-bounded or interrupt-on-update generation can truncate a tool call before its arguments are complete.
This couples computation time to speech updates, limiting useful work completed in advance.
We introduce Heartbeat Thinking, which handles updated input through new Thinker invocations while allowing ongoing invocations and dispatched tools to finish.
Periodic invocations read the latest partial transcript and a shared record of completed reasoning and tool results.
These results help resolve missing arguments and initiate dependent calls, allowing preparation to progress across speech updates.
Across SpeechFCEval, FDBench v3, and AudioCRAG, Heartbeat improves aggregate speculative tool coverage or retrieval reuse over a SHANKS-style windowed baseline, with benchmark-dependent effects on final task quality. On FDBench v3, it increases tool coverage by 16 percentage points and reduces mean final-response latency by 36.4\%. The results show that heartbeat thinking paradigm preserving ongoing computation during speech can turn listening time into reusable preparation for faster and more capable voice-agent responses.

}

\begin{document}
\makeanttitle
\raggedbottom
\suppressfloats[t]

\section{Introduction}

Voice agents use external tools to retrieve information and carry out requests that depend on actions beyond the conversation itself. For these tasks, producing a useful answer requires both model computation and external execution, which can leave users waiting after they finish speaking. Conventional cascaded pipelines often place this preparation after speech ends. Yet the information needed to begin may arrive much earlier: an agent can identify an order while the user is still describing a delivery concern. Prior work has shown that reasoning and tool use can begin on partial speech, allowing some of this preparation to take place during listening \citep{zhou2022online,chiang2025shanks,arora2025streamrag,hooper2026sia}. The opportunity is to reduce how much work remains when the agent must answer the complete request.

Early preparation is useful only when it produces work that can be reused for the full request. A tool call cannot be dispatched while its required arguments remain incomplete, even if the user has already provided enough information to construct them. SHANKS carries reasoning forward but bounds each generation episode by a speech chunk, which can truncate unfinished tool-call arguments \citep{chiang2025shanks}. SIA incorporates new events by interrupting ongoing generation; although dispatched tools are managed separately, a call may be interrupted before it is issued \citep{hooper2026sia}. Frequent updates can therefore limit the uninterrupted time available to form a complete call. The evolving request determines whether completed results remain useful: an early search may need revision when the user adds a budget constraint. Stream RAG explores parallel retrieval and checks whether intermediate results can serve the final request \citep{arora2025streamrag}. These considerations motivate a preparation policy that allows computation to finish across speech updates while letting later input determine which results still apply.

We therefore introduce \textbf{Heartbeat Thinking}, which handles new speech in subsequent invocations while allowing work already underway to finish (Figure~\ref{fig:heartbeat-framework}). During listening, periodic \emph{Thinker} invocations read the latest partial transcript and a shared record of completed reasoning and tool results. Independent invocations can overlap: one can finish constructing a tool call while another processes updated input. Requested tools execute asynchronously, and their returned results enter the shared record. Later invocations can use these results to resolve missing arguments and initiate dependent calls. Preparation can thus advance through successive reasoning and tool-use steps even when the transcript has not changed, without requiring each invocation to finish within an input-update window.

To make this accumulated preparation useful for the final answer, Heartbeat separates speculative preparation from response generation. At the detected end of speech, a \emph{Responder} receives the complete request and the completed preparation available at the handoff. It checks completed work against the full request, reuses applicable results, and performs the remaining calls. Unfinished speculative jobs are cancelled at this handoff, allowing the Responder to begin without waiting for their completion. This explicit exchange of reasoning records and tool results also allows the Thinker and Responder to use independently selected models.

Experiments on SpeechFCEval, FDBench v3, and AudioCRAG show higher aggregate speculative tool coverage or retrieval reuse than SHANKS-style Window Thinking. On FDBench v3, coverage increases from 34.67\% to 50.67\%, while mean final-response latency falls from 6.90 to 4.39 seconds (36.4\%), also 45.7\% below Pipecat, the fastest evaluated cascaded baseline. The coverage advantage persists across tested Responders. Task quality improves over Window Thinking on FDBench but decreases slightly on SpeechFCEval and AudioCRAG. Thus, computation spanning speech updates can increase reusable preparation and reduce final-response latency, while interpreting and using speculative results remains a separate challenge.

\section{Related Work}

\begin{figure}[!ht]
\centering
\includegraphics[width=\columnwidth]{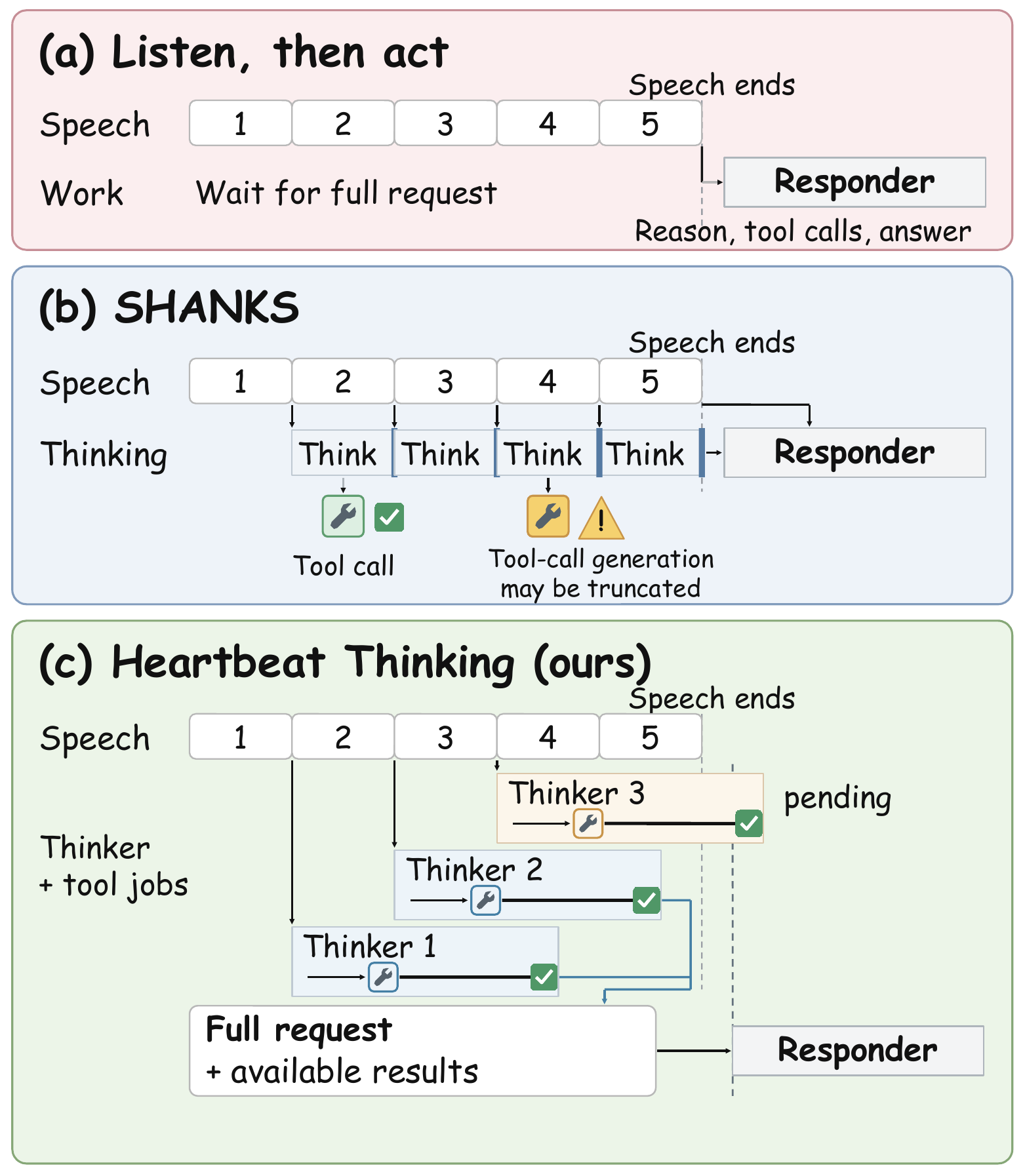}
\caption{Reasoning and tool use during speech. (a) Cascaded baselines defer reasoning and tool calls until speech ends. (b) SHANKS may truncate tool-call generation at chunk boundaries \citep{chiang2025shanks}. (c) Heartbeat overlaps Thinker invocations and tools across speech updates; the Responder reuses completed preparation without waiting for pending jobs.}
\label{fig:heartbeat-comparison}
\end{figure}

Early preparation spans prefix-based reasoning in LiveMind \citep{chen2024livemind}, response drafting in PredGen \citep{li2025predgen}, and maintained reasoning state in LTS-VoiceAgent \citep{zou2026ltsvoiceagent}. Figure~\ref{fig:heartbeat-comparison} contrasts when reasoning and tool use occur in cascaded pipelines, SHANKS, and Heartbeat. Decoupled systems exchange preparation through completed tool results in SHANKS \citep{chiang2025shanks}, streamed outputs in DuplexOmni \citep{huang2026duplexomni}, or model-internal state in AsyncReasoning \citep{yakushev2025asyncreasoning}. Heartbeat connects independently scheduled invocations through completed reasoning records and tool results. Appendix~\ref{app:related-work} discusses these interfaces and streaming computation policies in more detail.

\section{Heartbeat Thinking}
\label{sec:method}
\begin{figure*}[t]
\centering
\includegraphics[width=0.95\linewidth]{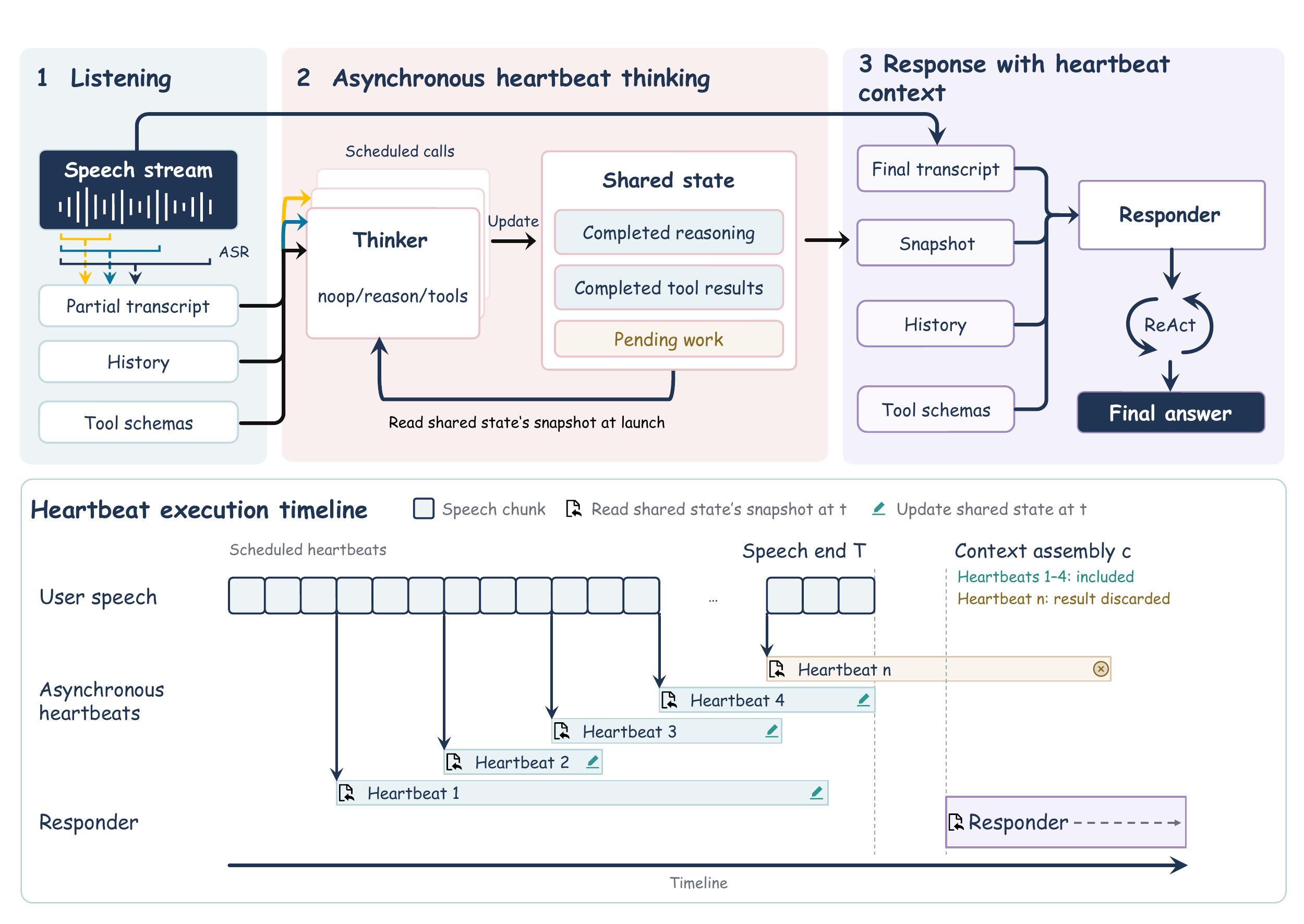}
\caption{Heartbeat Thinking architecture (top) and execution timeline (bottom). Thinker invocations and tools overlap during listening. At the VAD-triggered handoff $c$, unfinished speculative jobs are cancelled; the Responder uses completed preparation available at $c$ without waiting for pending work.}
\label{fig:heartbeat-framework}
\end{figure*}

Heartbeat receives streaming ASR transcripts, dialogue history, and tool schemas. A periodic \emph{Thinker} prepares reasoning and tool calls during listening; an endpoint-triggered \emph{Responder} answers the complete request. The roles can use different models. Shared state stores completed reasoning records, requested calls, and their returned results. Figure~\ref{fig:heartbeat-framework} shows the architecture and execution timeline. Appendix~\ref{app:method-details} provides implementation details and the output schema.

\paragraph{Overlapping preparation.}
Every $\Delta$ seconds, a new Thinker invocation reads the latest transcript and completed preparation. At most $K$ invocations run concurrently; a trigger is skipped when all slots are occupied. New speech and tool results enter subsequent invocations without interrupting ongoing generation. An invocation can take no action, return reasoning, or request tools with structured arguments. Completed reasoning is recorded before tools are dispatched asynchronously, so model and tool execution can overlap. Later invocations can use returned results to resolve missing arguments or initiate dependent calls, even without a new transcript update. The prompt directs the Thinker to avoid repeating pending or completed calls with identical names and arguments. Tool execution is queued separately from Thinker scheduling.

\begin{table*}[!t]
\centering
\small
\setlength{\tabcolsep}{5pt}
\renewcommand{\arraystretch}{1.05}
\begin{tabularx}{\textwidth}{@{}l*{7}{>{\centering\arraybackslash}X}@{}}
\toprule
\multirow{2}{*}[-0.5ex]{Framework}
 & \multicolumn{2}{c}{SpeechFCEval}
 & \multicolumn{3}{c}{FDBench v3}
 & \multicolumn{2}{c}{AudioCRAG} \\
\cmidrule(lr){2-3}\cmidrule(lr){4-6}\cmidrule(l){7-8}
 & Acc. & STC & Quality & Pass@1 & STC & Correct & STC \\
\midrule
Speech-to-Speech & 58.66 & 0.00 & 84.0 & 75.0 & 0.00 & 56.48 & 0.00 \\
Pipecat & 56.83 & 0.00 & 78.0 & 63.0 & 0.00 & 56.94 & 0.00 \\
LiveKit & 52.85 & 0.00 & 77.6 & 67.0 & 0.00 & 59.45 & 0.00 \\
Window Thinking (SHANKS) & 67.95 & 33.13 & 72.0 & 57.0 & 34.67 & 57.24 & 20.05 \\
\rowcolor{HBTableMethod}
\textbf{Heartbeat} & 66.85 & 43.69 & 77.0 & 65.0 & 50.67 & 55.49 & 62.42 \\
\bottomrule
\end{tabularx}
\caption{Main results (\%; higher is better). Speculative systems use Qwen3.6-35B-A3B as Thinker; all systems use DeepSeek-V4-Flash as Responder. Further task and model results are in Appendix~\ref{app:complete-results}.}
\label{tab:short-results}
\end{table*}

\paragraph{Response handoff.}
At the detected end of speech, the scheduler stops new invocations and cancels unfinished Thinker and speculative tool tasks, including queued calls. The Responder receives the final transcript, history, schemas, and a snapshot of completed preparation available at context assembly. Later speculative outputs are discarded. Through a ReAct process \citep{yao2023react}, it checks preparation against the complete request, rejects interpretations contradicted by later speech, and reuses applicable completed results. Required calls lacking available results are executed within the Responder's own process. It therefore begins without waiting for unfinished preparation. Appendix~\ref{app:boundary} details the availability boundary; Appendices~\ref{app:thinker-prompt}--\ref{app:responder-context} give the prompts.

\section{Evaluation}
\label{sec:experiments}

\begin{table}[!tb]
\centering
\small
\setlength{\tabcolsep}{2.6pt}
\setlength{\aboverulesep}{0.5pt}
\setlength{\belowrulesep}{1.5pt}
\renewcommand{\arraystretch}{1.05}
\begin{tabular}{@{}>{\columncolor{white}[0pt][\tabcolsep]}>{\raggedright\arraybackslash}m{\dimexpr\linewidth-80pt-4\tabcolsep\relax}>{\centering\arraybackslash}m{50pt}>{\columncolor{white}[\tabcolsep][0pt]\centering\arraybackslash}m{30pt}@{}}
\toprule
 \textbf{Framework} & \textbf{\mbox{Avg. Acc. $\uparrow$}} & \textbf{STC $\uparrow$} \\
\midrule

\multicolumn{3}{@{}>{\columncolor{HBTableGroup}[0pt][0pt]}l@{}}{\textit{Responder: DeepSeek-V4-Flash}} \\
Speech-to-Speech & 58.66 & 0.00 \\
Pipecat & 56.83 & 0.00 \\
LiveKit & 52.85 & 0.00 \\
Window Thinking (SHANKS) & 67.95 & 33.13 \\
\rowcolor{HBTableMethod}
\textbf{Heartbeat} & 66.85 & 43.69 \\
\midrule

\multicolumn{3}{@{}>{\columncolor{HBTableGroup}[0pt][0pt]}l@{}}{\textit{Responder: Qwen3.5-397B-A17B}} \\
Speech-to-Speech & 67.01 & 0.00 \\
Pipecat & 66.46 & 0.00 \\
LiveKit & 65.93 & 0.00 \\
Window Thinking (SHANKS) & 67.27 & 32.49 \\
\rowcolor{HBTableMethod}
\textbf{Heartbeat} & 69.70 & 41.20 \\
\midrule

\multicolumn{3}{@{}>{\columncolor{HBTableGroup}[0pt][0pt]}l@{}}{\textit{Responder: Qwen3.8-Flash-Next}} \\
Speech-to-Speech & 53.66 & 0.00 \\
Pipecat & 42.76 & 0.00 \\
LiveKit & 43.42 & 0.00 \\
Window Thinking (SHANKS) & 57.14 & 44.05 \\
\rowcolor{HBTableMethod}
\textbf{Heartbeat} & 64.30 & 55.27 \\
\bottomrule
\end{tabular}
\caption{SpeechFCEval accuracy and STC across Responders (\%), with Qwen3.6-35B-A3B as Thinker.}
\label{tab:responder-speechfc}
\end{table}

\begin{table}[!t]
\centering
\small
\setlength{\tabcolsep}{2.6pt}
\setlength{\aboverulesep}{0pt}
\setlength{\belowrulesep}{0pt}
\renewcommand{\arraystretch}{1.05}
\begin{tabular}{@{}>{\columncolor{white}[0pt][\tabcolsep]}>{\raggedright\arraybackslash}m{\dimexpr\linewidth-106pt-6\tabcolsep\relax}*{2}{>{\centering\arraybackslash}m{38pt}}>{\columncolor{white}[\tabcolsep][0pt]\centering\arraybackslash}m{30pt}@{}}
\toprule
 \textbf{Framework} & \textbf{Quality $\uparrow$} & \textbf{Pass@1 $\uparrow$} & \textbf{STC $\uparrow$} \\
\midrule

\multicolumn{4}{@{}>{\columncolor{HBTableGroup}[0pt][0pt]}l@{}}{\textit{Responder: DeepSeek-V4-Flash}} \\
Speech-to-Speech & 84.0 & 75.0 & 0.0 \\
Pipecat & 78.0 & 63.0 & 0.0 \\
LiveKit & 77.6 & 67.0 & 0.0 \\
Window Thinking (SHANKS) & 72.0 & 57.0 & 34.67 \\
\rowcolor{HBTableMethod}
\textbf{Heartbeat} & 77.0 & 65.0 & 50.67 \\
\midrule

\multicolumn{4}{@{}>{\columncolor{HBTableGroup}[0pt][0pt]}l@{}}{\textit{Responder: Qwen3.5-397B-A17B}} \\
Speech-to-Speech & 83.0 & 68.0 & 0.0 \\
Pipecat & 63.0 & 43.0 & 0.0 \\
LiveKit & 77.0 & 75.0 & 0.0 \\
Window Thinking (SHANKS) & 76.0 & 62.0 & 34.67 \\
\rowcolor{HBTableMethod}
\textbf{Heartbeat} & 77.0 & 61.0 & 45.33 \\
\midrule

\multicolumn{4}{@{}>{\columncolor{HBTableGroup}[0pt][0pt]}l@{}}{\textit{Responder: Qwen3.8-Flash-Next}} \\
Speech-to-Speech & 83.0 & 62.0 & 0.0 \\
Pipecat & 83.0 & 60.0 & 0.0 \\
LiveKit & 72.0 & 55.0 & 0.0 \\
Window Thinking (SHANKS) & 76.0 & 57.0 & 34.00 \\
\rowcolor{HBTableMethod}
\textbf{Heartbeat} & 74.0 & 59.0 & 50.67 \\
\bottomrule
\end{tabular}
\caption{FDBench v3 results across Responders (\%), with Qwen3.6-35B-A3B as Thinker.}
\label{tab:responder-fdbench}
\end{table}

\subsection{Setup and Metrics}
\label{sec:eval-setup}
\label{sec:primary-metrics}

We evaluate function calling on SpeechFCEval, intent following under disfluency on FDBench v3, and knowledge-grounded answering on AudioCRAG \citep{tongyi2026funaudio,lin2026fdbench,arora2025streamrag}. Baselines are Speech-to-Speech, Pipecat, LiveKit \citep{huggingface2024speechtospeech,daily2024pipecat,livekit2023agents}, and SHANKS-style Window Thinking. Window Thinking follows Heartbeat's update schedule but discards outputs arriving after their windows. We use Qwen3.6-35B-A3B as Thinker, DeepSeek-V4-Flash (0731) as Responder, $\Delta=500$\,ms, and $K=2$.

\emph{Speculative tool coverage} (STC) is the fraction of reference calls covered by completed matching speculative calls whose results enter the Responder context without the Responder repeating those calls. On AudioCRAG, it instead measures the fraction of samples reusing completed speculative results without additional retrieval. Task quality uses SpeechFCEval function-name/argument accuracy, FDBench Quality and Pass@1, and AudioCRAG correct-answer rate. Appendix~\ref{app:benchmarks} gives dataset sizes, matching and aggregation rules, and scoring protocols.

\subsection{Results}
\begin{figure}[t]
\centering
\includegraphics[width=0.75\linewidth]{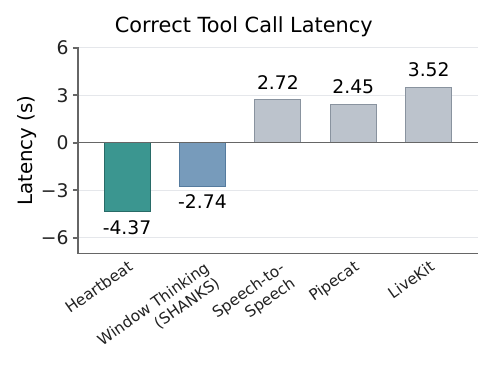}
\caption{Mean Correct Tool Call Latency on FDBench v3 relative to speech end, measured at initiation of the first tool call matching the final request. Negative values indicate events occurring before speech end. Numerical results are reported in Table~\ref{tab:fdbench-latency-summary}.}
\label{fig:fdbench-latency}
\end{figure}

\begin{figure}[t]
\centering
\includegraphics[width=0.75\linewidth]{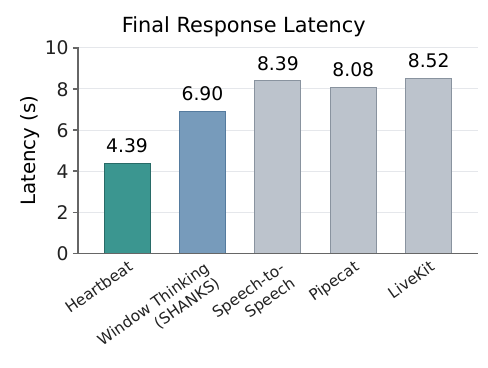}
\caption{Mean Final Response Latency on FDBench v3 relative to speech end, measured at the onset of the final spoken response. Details are reported in Table~\ref{tab:fdbench-latency-summary}.}
\label{fig:fdbench-final-latency}
\end{figure}

Table~\ref{tab:short-results} shows higher STC for Heartbeat than Window Thinking on all three benchmarks: gains of 10.56, 16.00, and 42.37 percentage points on SpeechFCEval, FDBench, and AudioCRAG. On FDBench, Quality rises from 72.0\% to 77.0\% and Pass@1 from 57.0\% to 65.0\%, although Pass@1 remains below Speech-to-Speech and LiveKit. Final task performance decreases on SpeechFCEval (67.95\% to 66.85\%) and AudioCRAG (57.24\% to 55.49\%). More retained preparation therefore does not consistently yield better final task performance.

\paragraph{Latency.}
\label{sec:latency}
Figures~\ref{fig:fdbench-latency} and~\ref{fig:fdbench-final-latency} show mean FDBench latencies relative to speech end. Correct Tool Call Latency measures initiation of the first matching execution, conditional on a matching call. Heartbeat starts at $-4.37$\,s versus $-2.74$\,s for Window Thinking and 2.45--3.52\,s for cascaded baselines. Final Response Latency measures the onset of the final spoken answer after required tool calls, excluding earlier acknowledgments. Heartbeat reduces it from Window Thinking's 6.90 to 4.39\,s (36.4\%), also 45.7\% below Pipecat, the fastest evaluated cascaded baseline. Appendix~\ref{app:fdbench-latency-protocol} gives the protocol and numerical results.

\paragraph{Additional analyses.}
\label{sec:analysis}
Heartbeat retains higher STC across all three tested Responders (Tables~\ref{tab:responder-speechfc} and~\ref{tab:responder-fdbench}; Appendix~\ref{sec:ablations}). Later invocations can incorporate late constraints and explicit intent revisions before speech ends (Appendix~\ref{app:speculative-details}). Yet 4.0\% of FDBench recordings yield answers using erroneous or outdated results despite correct tool-call arguments, distinguishing call correctness from result use.

\section{Conclusion}
Heartbeat Thinking overlaps speculative computation across speech updates and passes completed preparation to a separate Responder. Across three benchmarks, it retains more reusable preparation than a SHANKS-style windowed baseline; on FDBench, it also reduces final-response latency. Task-quality changes vary by benchmark, highlighting the importance of interpreting and using speculative results correctly.

\section*{Limitations}
\label{sec:limitations}
Heartbeat uses a fixed scheduling interval and a bounded number of concurrent Thinker invocations. We have not explored adaptive scheduling based on changes in the partial transcript or the completion of tool calls. Such scheduling could reduce redundant inference when little new information is available. The Responder also receives a snapshot of completed preparation at context assembly, excluding tool results returned afterward. Future work could explore incremental delivery of late results, particularly for tools with variable execution times. Finally, our latency analysis focuses on FDBench v3, and further evaluation across different tool workloads would help characterize the trade-off between speculative coverage, inference cost, and response latency.

\section*{Ethical Considerations}
Speech and dialogue traces may contain personal information; applicable consent and access controls must govern their collection and use. The SpeechFCEval dataset card declares Apache-2.0 and asks users to respect the terms of its upstream sources.\footnote{\url{https://huggingface.co/datasets/FunAudioLLM/SpeechFCEval}} The FDBench repository and the original CRAG text-data repository publish CC BY-NC 4.0 licenses.\footnote{\url{https://github.com/DanielLin94144/Full-Duplex-Bench/blob/main/LICENSE}; \url{https://github.com/facebookresearch/CRAG/blob/main/LICENSE}} These upstream terms do not establish permission to redistribute derived audio or identifiable dialogue traces.

\bibliographystyle{acl_natbib}
\bibliography{references}

\appendix
\section{Detailed Framework}\label{app:method-details}

\subsection{Task and framework overview}
Let $x_{[0,t]}$ denote speech observed through time $t$, $H$ the dialogue history including tool calls and results from previous turns, and $\mathcal{T}$ the available tool schemas. The current utterance ends at $T$. The task is to answer the complete request and execute its required tool calls using $H$, $\mathcal{T}$, and $x_{[0,T]}$. Streaming automatic speech recognition (ASR) provides partial transcripts during listening and a final transcript of the complete utterance.

Heartbeat separates preparation and response generation into two language-model roles (Figure~\ref{fig:heartbeat-framework}). The \emph{Thinker} is invoked periodically during listening to produce provisional reasoning and tool-call requests from partial transcripts. At the detected end of speech, speculative preparation stops and a separate \emph{Responder} is triggered. It uses the final transcript and the preparation already available to complete the remaining reasoning, tool calls, and final answer.

Both roles receive $H$ and $\mathcal{T}$ and may use different models. A shared state stores current-turn preparation: completed Thinker outputs, their tool-call requests, and the associated results as they return. Tools execute asynchronously, independently of the model invocations.

\subsection{Periodic asynchronous Thinker}
Heartbeats occur at a fixed interval $\Delta$ during listening, with at most $K$ Thinker invocations running concurrently. A trigger is skipped rather than queued when $K$ invocations are already active. Otherwise, a new invocation starts without interrupting earlier ones.

An invocation started at $t_k$ receives $H$, $\mathcal{T}$, the latest partial transcript, completed outputs from earlier Thinker invocations in the current turn, and results from their tool calls available at $t_k$. New speech and newly returned tool results enter subsequent invocations rather than being injected into ongoing generation.

The Thinker can produce three forms of output: no additional action, reasoning only, or reasoning with tool requests. These are encoded as \texttt{noop}, \texttt{think} without tool calls, and \texttt{think} with tool calls, respectively. A \texttt{noop} indicates that the Thinker identifies no need for additional reasoning or tool requests given the current input and available preparation. Tool requests specify tool names and structured arguments once the necessary information is available. The prompt and output schema are given in Appendix~\ref{app:thinker-prompt}.

Once a \texttt{think} output is complete, its reasoning record and any tool-call requests are written to the shared state, after which the requested tools are dispatched asynchronously. Since one invocation may request multiple calls, tool concurrency is controlled separately from Thinker concurrency: dispatch imposes no concurrency limit, while the execution backend queues calls for execution. A recorded call is \emph{pending} until its result returns; the result is then linked to the request by its unique tool-call ID and stored alongside it.

These records allow subsequent Thinker invocations to distinguish work already requested from results available for further reasoning. To avoid redundant calls, the prompt instructs each invocation not to repeat pending or completed calls with the same tool name and complete arguments. Returned results can also supply new information for dependent calls, allowing preparation to advance even when the partial transcript is unchanged.

\subsection{Endpoint-triggered Responder}
The voice activity detection (VAD) module triggers the transition to response generation at the detected end of speech. At this point, the scheduler stops launching new Thinker invocations and cancels unfinished Thinker and speculative tool tasks, including queued calls. The Responder receives the final transcript, the shared history $H$, the tool schemas $\mathcal{T}$, and the preparation available at this transition.

Let $c$ denote the VAD-triggered context-assembly time and $a_j$ the availability time of preparation output $r_j$. The preparation snapshot is
\begin{equation}
\mathcal{R}_c=\{r_j:r_j\text{ is complete and }a_j\leq c\}.
\label{eq:availability}
\end{equation}
The snapshot contains completed Thinker records and returned tool results. A completed record may include tool-call requests whose results are not yet available. Outputs returned by speculative jobs after $c$ are discarded. Appendix~\ref{app:boundary} illustrates the boundary cases.

The Responder follows a ReAct process \citep{yao2023react}. It checks the preparation against the complete request and disregards interpretations contradicted by later speech. Its prompt directs it to reuse available results from completed calls that match the required operations in tool name and complete arguments. Required calls without available results, including those pending at the handoff, are issued by the Responder as part of its own ReAct process. Its system prompt and instructions for reading the Heartbeat context appear in Appendices~\ref{app:responder-prompt} and~\ref{app:responder-context}. The answer can then be passed to text-to-speech, and the dialogue and retained tool-call records and results are carried forward to subsequent turns.

\section{Benchmark Setup}
\label{app:benchmarks}

For the main task-quality comparisons, the DeepSeek-V4-Flash Responder uses the 0731 version. The cascaded pipelines combine Qwen3-ASR with a text reasoning model. All experiments use NVIDIA H200 GPUs.

\paragraph{Coverage and accuracy aggregation.}
For SpeechFCEval and FDBench, STC measures the fraction of reference calls covered by completed speculative calls whose results enter the Responder context without repetition by the Responder. Reference matching follows benchmark-specific tool-name and argument rules; repetition requires identical tool names and complete arguments. SpeechFCEval averages per-sample coverage, whereas FDBench aggregates over reference calls and checks repetitions across each recording. AudioCRAG STC measures the fraction of samples reusing completed speculative results without additional retrieval. SpeechFCEval task accuracy requires correct function names and arguments and is averaged equally over its five task groups.

\subsection{SpeechFCEval}
SpeechFCEval combines speech-based tasks from ACEBench and BFCL with SmartInteract \citep{tongyi2026funaudio}. It covers single and parallel function calling across synthesized and human speech. We evaluate the full test set.

The test set contains 1,914 samples: 208 and 88 in Speech-ACEBench single and parallel calling, 578 and 372 in Speech-BFCL single and parallel calling, and 668 in SmartInteract. All samples remain in the aggregate, with runtime failures contributing zero. Overall accuracy is averaged equally across the five task groups, while Overall STC is averaged across individual samples.

The model comparisons use a 500\,ms heartbeat interval and at most two concurrent Thinker invocations. Thinking is disabled for the two Qwen Responders.

\subsection{FDBench v3}
FDBench v3 tests whether voice agents follow spoken intent under disfluency \citep{lin2026fdbench}. It evaluates turn taking, tool selection, argument accuracy, response quality, and task success. Spoken responses are transcribed for evaluation, and task performance is assessed from the executed calls and resulting answers.

FDBench v3 contains 100 recordings covering 79 scenarios and 12 speakers. Spoken responses are transcribed with NVIDIA Parakeet TDT 0.6B v2.

We use DeepSeek-V4-Flash with maximum reasoning effort as the judge for FDBench v3. Quality follows the benchmark author's answer-quality assessment. For Heartbeat, Pass@1 requires completed speculative results and Responder calls jointly to fulfill the task without incorrect or extra Responder calls.

The cascaded baselines use the same judge for argument accuracy, response quality, and Pass@1. Turn taking and tool selection follow the benchmark's original evaluation.

\paragraph{Latency configuration.}
\label{app:fdbench-latency-runtime}
The latency evaluation uses Qwen3.6-35B-A3B as the Thinker and DeepSeek-V4-Flash as the Responder, with thinking disabled for both models. Table~\ref{tab:fdbench-latency-runtime} lists the speech components and timing settings.

\begin{table}[t]
\centering
\small
\renewcommand{\arraystretch}{1.12}
\setlength{\tabcolsep}{5pt}
\renewcommand{\tabularxcolumn}[1]{m{#1}}
\begin{tabularx}{\columnwidth}{@{}>{\raggedright\arraybackslash}m{0.42\columnwidth}>{\raggedright\arraybackslash}X@{}}
\toprule
Parameter & Value \\
\midrule
ASR & Qwen3-ASR-1.7B \\
TTS & Qwen3-TTS-12Hz-\allowbreak1.7B-Base \\
\addlinespace[4pt]
ASR update block & 500\,ms \\
VAD minimum speech & 200\,ms \\
VAD minimum silence & 800\,ms \\
VAD padding & 500\,ms \\
\addlinespace[4pt]
Heartbeat interval & 500\,ms \\
Max. heartbeat concurrency & 2 \\
\bottomrule
\end{tabularx}
\caption{Speech pipeline and heartbeat scheduling settings for the FDBench v3 latency evaluation.}
\label{tab:fdbench-latency-runtime}
\end{table}

\paragraph{Latency protocol.}
\label{app:fdbench-latency-protocol}
Latencies are measured relative to each input turn's aligned speech endpoint, excluding trailing silence.

Final Response Latency measures the onset of the final spoken reply after the last tool call, or the onset of the reply when no tool call is required. Correct Tool Call uses the recorded initiation time of the first eligible call matching the turn's reference. Matching uses exact tool names with the benchmark's string normalization and dynamic-argument handling. Speculative calls count only after completion and inclusion in the Responder context.

\paragraph{Latency aggregation.}
We first average event offsets within each recording, then give recordings equal weight. Negative offsets are retained. Correct Tool Call Latency is conditional on a matching call. Table~\ref{tab:fdbench-latency-summary} reports the system-level means.

\begin{table}[t]
\centering
\small
\setlength{\tabcolsep}{2.6pt}
\begin{tabular}{@{}>{\raggedright\arraybackslash}m{\dimexpr\linewidth-132pt-4\tabcolsep\relax}*{2}{>{\centering\arraybackslash}m{66pt}}@{}}
\toprule
System & \shortstack{Correct\\Tool Call} & \shortstack{Final\\Response} \\
\midrule
\textbf{Heartbeat} & $-4.37$ & 4.39 \\
\shortstack[l]{Window Thinking\\(SHANKS)} & $-2.74$ & 6.90 \\
Speech-to-Speech & $2.72$ & 8.39 \\
Pipecat & $2.45$ & 8.08 \\
LiveKit & $3.52$ & 8.52 \\
\bottomrule
\end{tabular}
\caption{Mean event latencies on FDBench v3 (seconds), measured relative to speech end. Negative values indicate events occurring before speech end.}
\label{tab:fdbench-latency-summary}
\end{table}

\subsection{AudioCRAG}
AudioCRAG evaluates knowledge-grounded spoken answering \citep{arora2025streamrag}.

The AudioCRAG evaluation set contains 1,312 samples. Heartbeat uses a 500\,ms interval and at most two concurrent Thinker invocations.

\section{Complete Results}\label{app:complete-results}
\newcommand{\hbmodelwidth}{\dimexpr\linewidth-78pt-138pt-54pt-33pt-14\tabcolsep\relax}
\newcommand{\hbdescriptionwidth}{\dimexpr78pt+\hbmodelwidth+2\tabcolsep\relax}
\newcommand{\hbtaskheader}{%
& \multicolumn{2}{c}{\textbf{Speech-ACEBench}}
& \multicolumn{2}{c}{\textbf{Speech-BFCL}}
& \textbf{SmartInteract} & \\
}

\begin{table*}[!t]
\centering
\small
\setlength{\tabcolsep}{2.6pt}
\setlength{\aboverulesep}{0pt}
\setlength{\belowrulesep}{0pt}
\renewcommand{\arraystretch}{1.05}
\captionsetup[subtable]{position=top,font=10pt,justification=raggedright,singlelinecheck=false}
\begin{subtable}{\linewidth}
\caption{Function-calling accuracy (Acc.; $\uparrow$) across the five task groups.}
\label{tab:speechfc-cascaded}
\begin{tabular}{@{}>{\columncolor{white}[0pt][\tabcolsep]}m{78pt}>{\raggedright\arraybackslash}m{\hbmodelwidth}*{4}{>{\centering\arraybackslash}m{34.5pt}}>{\centering\arraybackslash}m{54pt}>{\columncolor{white}[\tabcolsep][0pt]\centering\arraybackslash}m{33pt}@{}}
\toprule
 & \hbtaskheader
\multirow{-2}{*}{\textbf{Framework}} & \multirow{-2}{*}{\textbf{Thinker}} & Single & Parallel & Single & Parallel & Single & \multirow{-2}{*}{\textbf{Overall}} \\
\midrule
\multicolumn{8}{@{}l}{\textit{Cascaded}} \\
Speech-to-Speech & -- & 49.00 & 25.00 & 91.00 & 60.48 & 67.81 & 58.66 \\
Pipecat & -- & 55.80 & 20.50 & 89.79 & 56.99 & 61.08 & 56.83 \\
LiveKit & -- & 44.20 & 13.60 & 89.27 & 58.06 & 59.13 & 52.85 \\
\midrule
\mbox{Window Thinking}\newline(SHANKS) & Qwen3.6-35B-A3B & 63.94 & 44.32 & 86.16 & 71.51 & 73.80 & 67.95 \\
\midrule
\rowcolor{HBTableMethod}
 & Qwen3.6-35B-A3B & 57.21 & 37.50 & 86.68 & 76.08 & 76.80 & 66.85 \\
\rowcolor{HBTableMethod}
 & Qwen3.5-9B & 46.63 & 21.59 & 82.87 & 72.58 & 71.56 & 59.05 \\
\rowcolor{HBTableMethod}
 & GLM-4.7-Flash & 45.19 & 22.73 & 83.91 & 71.77 & 73.65 & 59.45 \\
\rowcolor{HBTableMethod}
\multirow{-4}{*}{\textbf{Heartbeat}} & Gemma-4-26B-A4B-it & 62.50 & 37.50 & 87.20 & 76.08 & 79.64 & 68.58 \\
\bottomrule
\end{tabular}
\end{subtable}

\vspace{8pt}
\begin{subtable}{\linewidth}
\caption{Speculative tool coverage (STC; $\uparrow$) across the five task groups.}
\label{tab:speechfc-stc}
\begin{tabular}{@{}>{\columncolor{white}[0pt][\tabcolsep]}m{78pt}>{\raggedright\arraybackslash}m{\hbmodelwidth}*{4}{>{\centering\arraybackslash}m{34.5pt}}>{\centering\arraybackslash}m{54pt}>{\columncolor{white}[\tabcolsep][0pt]\centering\arraybackslash}m{33pt}@{}}
\toprule
 & \hbtaskheader
\multirow{-2}{*}{\textbf{Framework}} & \multirow{-2}{*}{\textbf{Thinker}} & Single & Parallel & Single & Parallel & Single & \multirow{-2}{*}{\textbf{Overall}} \\
\midrule
\multicolumn{2}{@{}l}{Cascaded} & 0.00 & 0.00 & 0.00 & 0.00 & 0.00 & 0.00 \\
\midrule
\mbox{Window Thinking}\newline(SHANKS) & Qwen3.6-35B-A3B & 28.37 & 0.00 & 29.24 & 36.34 & 40.57 & 33.13 \\
\midrule
\rowcolor{HBTableMethod}
 & Qwen3.6-35B-A3B & 42.31 & 36.78 & 42.21 & 58.03 & 38.32 & 43.69 \\
\rowcolor{HBTableMethod}
 & Qwen3.5-9B & 28.85 & 23.66 & 35.81 & 53.51 & 25.00 & 34.16 \\
\rowcolor{HBTableMethod}
 & GLM-4.7-Flash & 35.10 & 20.87 & 39.97 & 49.67 & 35.93 & 39.04 \\
\rowcolor{HBTableMethod}
\multirow{-4}{*}{\textbf{Heartbeat}} & Gemma-4-26B-A4B-it & 49.04 & 38.98 & 42.73 & 58.91 & 32.19 & 42.71 \\
\bottomrule
\end{tabular}
\end{subtable}
\caption{SpeechFCEval results (\%) with DeepSeek-V4-Flash as the Responder. (a) Function-calling accuracy and (b) speculative tool coverage (STC).}
\label{tab:speechfc}
\end{table*}

\begin{table*}[!t]
\centering
\small
\setlength{\tabcolsep}{4pt}
\setlength{\aboverulesep}{2pt}
\setlength{\belowrulesep}{3pt}
\renewcommand{\arraystretch}{1.05}
\begin{tabular}{@{}>{\columncolor{white}[0pt][\tabcolsep]\raggedright\arraybackslash}m{160pt}>{\raggedright\arraybackslash}m{120pt}*{2}{>{\centering\arraybackslash}m{\dimexpr(\linewidth-280pt-8\tabcolsep)/3\relax}}>{\columncolor{white}[\tabcolsep][0pt]\centering\arraybackslash}m{\dimexpr(\linewidth-280pt-8\tabcolsep)/3\relax}@{}}
\toprule
 \textbf{Framework} & \textbf{Thinker} & \textbf{Quality} & \textbf{Pass@1} & \textbf{STC} \\
\midrule
\multicolumn{5}{@{}l}{\textit{Cascaded}} \\
Speech-to-Speech & -- & 84.0 & 75.0 & 0.0 \\
Pipecat & -- & 78.0 & 63.0 & 0.0 \\
LiveKit & -- & 77.6 & 67.0 & 0.0 \\
\midrule
Window Thinking (SHANKS) & Qwen3.6-35B-A3B & 72.0 & 57.0 & 34.67 \\
\midrule
\rowcolor{HBTableMethod}
 & Qwen3.6-35B-A3B & 77.0 & 65.0 & 50.67 \\
\rowcolor{HBTableMethod}
 & Qwen3.5-9B & 80.0 & 57.0 & 39.33 \\
\rowcolor{HBTableMethod}
 & GLM-4.7-Flash & 84.0 & 55.0 & 43.33 \\
\rowcolor{HBTableMethod}
\multirow{-4}{*}{\textbf{Heartbeat}} & Gemma-4-26B-A4B-it & 80.0 & 67.0 & 52.00 \\
\bottomrule
\end{tabular}

\caption{FDBench v3 task performance and speculative tool coverage (STC; \%) with DeepSeek-V4-Flash as the Responder.}
\label{tab:fdbench}
\end{table*}

\newcommand{\cragcell}[2]{#2}
\newcommand{\cragmodelwidth}{\dimexpr\linewidth-78pt-54pt-54pt-78pt-54pt-10\tabcolsep\relax}
\newcommand{\cragoutcomewidth}{\dimexpr(\linewidth-78pt-\cragmodelwidth-8\tabcolsep)/3\relax}
\newcommand{\cragcoveragewidth}{\dimexpr\linewidth-78pt-\cragmodelwidth-4\tabcolsep\relax}

\begin{table*}[t]
\centering
\small
\setlength{\tabcolsep}{2.6pt}
\setlength{\aboverulesep}{0pt}
\setlength{\belowrulesep}{0pt}
\renewcommand{\arraystretch}{1.05}
\begin{tabular}{@{}>{\columncolor{white}[0pt][\tabcolsep]}m{78pt}>{\raggedright\arraybackslash}m{\cragmodelwidth}*{2}{>{\centering\arraybackslash}m{54pt}}>{\centering\arraybackslash}m{78pt}>{\columncolor{white}[\tabcolsep][0pt]\centering\arraybackslash}m{54pt}@{}}
\toprule
 \textbf{Framework} & \textbf{Thinker} & \textbf{Correct $\uparrow$} & \textbf{Missing} & \textbf{Hallucination $\downarrow$} & \textbf{STC $\uparrow$} \\
\midrule
\multicolumn{6}{@{}l}{\textit{Cascaded}} \\
Speech-to-Speech & -- & \cragcell{741}{56.48} & \cragcell{157}{11.97} & \cragcell{414}{31.55} & 0.00 \\
Pipecat & -- & \cragcell{747}{56.94} & \cragcell{157}{11.97} & \cragcell{408}{31.10} & 0.00 \\
LiveKit & -- & \cragcell{780}{59.45} & \cragcell{169}{12.88} & \cragcell{363}{27.67} & 0.00 \\
\midrule
\mbox{Window Thinking}\newline(SHANKS) & Qwen3.6-35B-A3B & \cragcell{751}{57.24} & \cragcell{188}{14.33} & \cragcell{373}{28.43} & 20.05 \\
\midrule
\rowcolor{HBTableMethod}
 & Qwen3.6-35B-A3B & \cragcell{728}{55.49} & \cragcell{221}{16.84} & \cragcell{363}{27.67} & 62.42 \\
\rowcolor{HBTableMethod}
 & Qwen3.5-9B & \cragcell{712}{54.27} & \cragcell{240}{18.29} & \cragcell{360}{27.44} & 60.06 \\
\rowcolor{HBTableMethod}
 & GLM-4.7-Flash & \cragcell{686}{52.29} & \cragcell{254}{19.36} & \cragcell{372}{28.35} & 63.57 \\
\rowcolor{HBTableMethod}
\multirow{-4}{*}{\textbf{Heartbeat}} & Gemma-4-26B-A4B-it & \cragcell{733}{55.87} & \cragcell{218}{16.62} & \cragcell{361}{27.52} & 74.70 \\
\bottomrule
\end{tabular}
\caption{AudioCRAG answer outcomes and retrieval reuse (\%) with DeepSeek-V4-Flash as the Responder. Here, STC measures the share of samples with completed speculative results included in the Responder's context and no additional retrieval by the Responder.}
\label{tab:audiocrag}
\end{table*}

\subsection{Task Results}

\paragraph{SpeechFCEval.}
Heartbeat achieves 43.69\% STC, compared with 33.13\% for the SHANKS-style Window Thinking baseline, an improvement of 10.56 percentage points. Across the five task groups reported in Table~\ref{tab:speechfc}, coverage increases in four, with Speech-BFCL parallel calling improving from 36.34\% to 58.03\%. SmartInteract is the exception, where coverage decreases slightly from 40.57\% to 38.32\%. Despite achieving higher coverage, Heartbeat reaches 66.85\% overall function-calling accuracy, slightly below Window Thinking at 67.95\%, although both outperform the cascaded baselines. This distinction indicates that retaining more speculative tool results does not necessarily improve function-calling accuracy.

\paragraph{FDBench v3.}
Heartbeat achieves 50.67\% STC, outperforming Window Thinking's 34.67\% by 16.00 percentage points. Unlike SpeechFCEval, its task-quality scores also improve: Table~\ref{tab:fdbench} records an increase in response quality from 72.0\% to 77.0\% and in Pass@1 from 57.0\% to 65.0\%. These results show that Heartbeat retains more speculative tool work while achieving better task success than the windowed baseline under spoken disfluency. However, its Pass@1 remains below those of Speech-to-Speech and LiveKit.

\paragraph{AudioCRAG.}
On AudioCRAG, Heartbeat reaches 62.42\% STC but achieves a correct-answer rate of 55.49\%. The cascaded baselines listed in Table~\ref{tab:audiocrag} achieve correct-answer rates between 56.48\% and 59.45\%. Window Thinking achieves 20.05\% STC and a 57.24\% correct-answer rate. Heartbeat's missing-answer rate increases from Window Thinking's 14.33\% to 16.84\%. This contrast further illustrates that higher speculative retrieval reuse does not guarantee better answer quality. The benefit of early retrieval depends not only on retaining completed results but also on whether the Responder can use them correctly.

\subsection{Model Comparisons}

\label{sec:ablations}
We vary the Thinker and Responder separately to test preparation and reuse across models.

\paragraph{Thinker choice.}
All four Thinkers achieve nonzero STC on all three benchmarks (Tables~\ref{tab:speechfc}--\ref{tab:audiocrag}). Higher coverage does not consistently yield better task performance: on SpeechFCEval, Gemma achieves higher accuracy than the default Thinker (68.58\% vs.\ 66.85\%) despite slightly lower STC (42.71\% vs.\ 43.69\%).

\paragraph{Responder choice.}
Heartbeat retains higher STC than Window Thinking with all three Responders on all three benchmarks, including FDBench (Table~\ref{tab:responder-fdbench}). However, model rankings differ by metric: on SpeechFCEval, Qwen3.8-Flash-Next achieves higher STC than Qwen3.5-397B-A17B (55.27\% vs.\ 41.20\%) but lower accuracy (64.30\% vs.\ 69.70\%).

\begin{table}[!tb]
\centering
\small
\setlength{\tabcolsep}{1.8pt}
\setlength{\aboverulesep}{0.5pt}
\setlength{\belowrulesep}{1.5pt}
\renewcommand{\arraystretch}{1.16}
\begin{tabular}{@{}>{\columncolor{white}[0pt][\tabcolsep]}>{\raggedright\arraybackslash}m{\dimexpr\linewidth-133pt-8\tabcolsep\relax}>{\centering\arraybackslash}m{42pt}>{\centering\arraybackslash}m{27pt}>{\centering\arraybackslash}m{37pt}>{\columncolor{white}[\tabcolsep][0pt]\centering\arraybackslash}m{27pt}@{}}
\toprule
 \textbf{Framework} & \textbf{\mbox{Correct $\uparrow$}} & \textbf{Miss.} & \textbf{\mbox{Hallu. $\downarrow$}} & \textbf{STC $\uparrow$} \\
\midrule

\multicolumn{5}{@{}>{\columncolor{HBTableGroup}[0pt][0pt]}l@{}}{\textit{Responder: DeepSeek-V4-Flash}} \\
Speech-to-Speech & 56.48 & 11.97 & 31.55 & 0.00 \\
Pipecat & 56.94 & 11.97 & 31.10 & 0.00 \\
LiveKit & 59.45 & 12.88 & 27.67 & 0.00 \\
Window Thinking (SHANKS) & 57.24 & 14.33 & 28.43 & 20.05 \\
\rowcolor{HBTableMethod}
\textbf{Heartbeat} & 55.49 & 16.84 & 27.67 & 62.42 \\
\midrule

\multicolumn{5}{@{}>{\columncolor{HBTableGroup}[0pt][0pt]}l@{}}{\textit{Responder: Qwen3.5-397B-A17B}} \\
Speech-to-Speech & 52.29 & 8.00 & 39.71 & 0.00 \\
Pipecat & 55.56 & 6.25 & 38.19 & 0.00 \\
LiveKit & 57.16 & 7.32 & 35.52 & 0.00 \\
Window Thinking (SHANKS) & 55.79 & 9.76 & 34.45 & 20.43 \\
\rowcolor{HBTableMethod}
\textbf{Heartbeat} & 55.95 & 8.16 & 35.90 & 66.46 \\
\midrule

\multicolumn{5}{@{}>{\columncolor{HBTableGroup}[0pt][0pt]}l@{}}{\textit{Responder: Qwen3.8-Flash-Next}} \\
Speech-to-Speech & 56.86 & 11.20 & 31.94 & 0.00 \\
Pipecat & 54.27 & 11.97 & 33.77 & 0.00 \\
LiveKit & 58.00 & 10.59 & 31.40 & 0.00 \\
Window Thinking (SHANKS) & 53.35 & 16.01 & 30.64 & 11.05 \\
\rowcolor{HBTableMethod}
\textbf{Heartbeat} & 56.17 & 15.17 & 28.66 & 40.70 \\
\bottomrule
\end{tabular}
\caption{AudioCRAG answer outcomes and retrieval reuse across Responder models (\%), with Qwen3.6-35B-A3B as the Thinker. The highest reuse rate and correct-answer rate occur with different Responders.}
\label{tab:responder-audiocrag}
\end{table}

\section{Speculative Tool-Use Analysis Details}
\label{app:speculative-details}

\paragraph{Grouping and success criteria.}
We group recordings by when the target entity and all constraints of the final request have been fully spoken. Early information arrives within the first 70\% of the effective input interval; late information arrives within the final 30\%. Effective input duration excludes leading and trailing silence. Explicit intent revisions form a separate group.

For the early- and late-information groups, we measure the fraction of recordings in which a call with required arguments consistent with the final request is initiated before speech end. For the revision group, correction succeeds when an earlier call inconsistent with the final request is followed by a corrected call before speech end. These initiation-based measures differ from STC, which requires completed results to be included in the Responder context without response-stage repetition.

\paragraph{Inconsistent result reuse.}
In these failures, a correct-argument tool request coexists with a final answer that uses erroneous or outdated speculative results.

For example, the system adds a \$99.99 coffee maker to the cart before hearing the complete \$50 budget, then retrieves a \$40 product. The answer claims that the cheaper product was added but reports the earlier \$99.99 cart total, mixing the two results. The corrected search does not undo the cart action.

\subsection{Timing and Case Studies}
\label{sec:speculative_analysis}

\begin{figure*}[t]
\centering
\includegraphics[width=\textwidth]{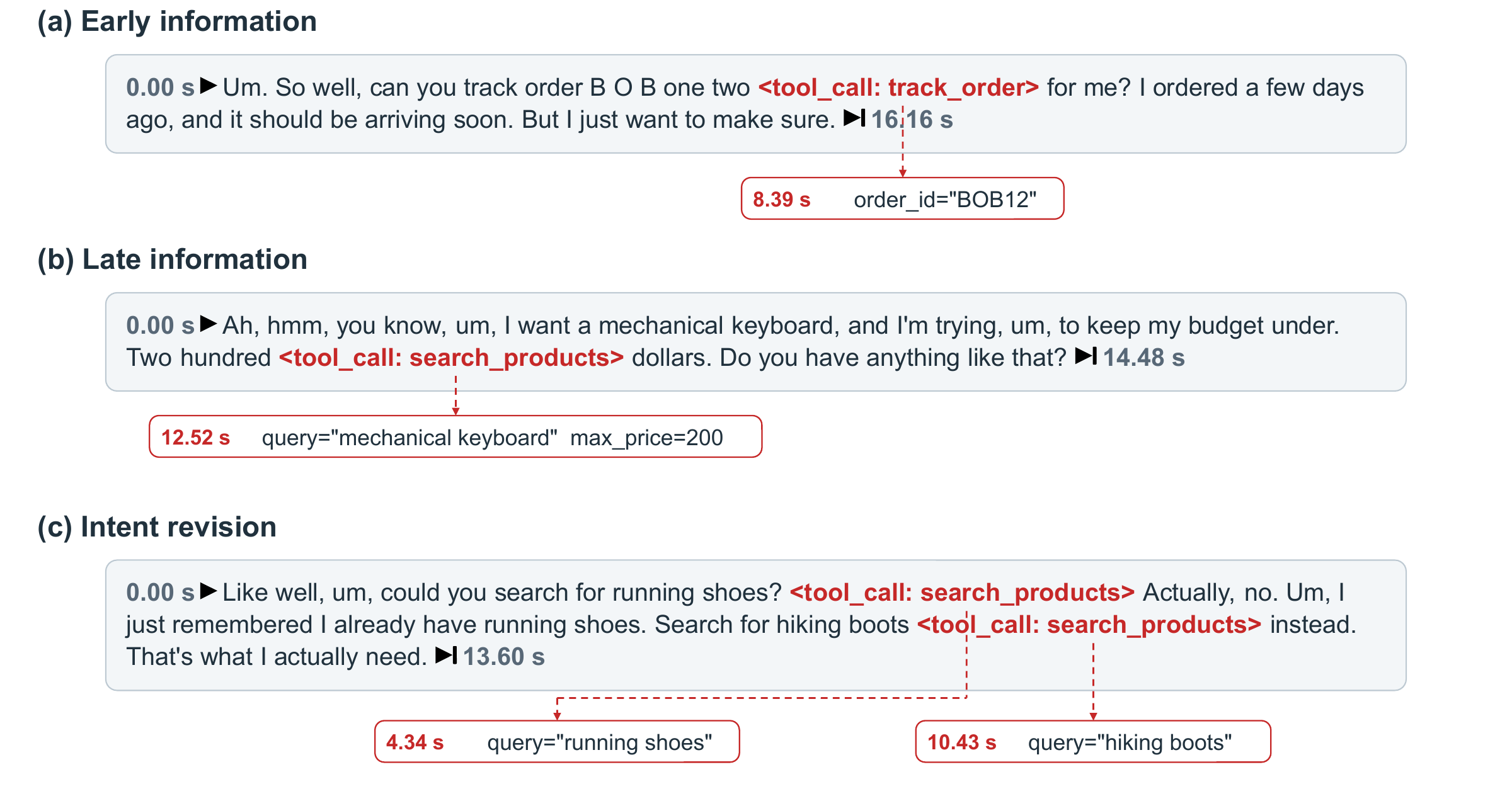}
\caption{Speculative tool calls during listening on FDBench v3. Red markers indicate the last words available to the Thinker when each invocation began; callouts give tool-request initiation times and arguments. Times are relative to audio onset. Word alignment is approximate, and ASR text is normalized for display.}\label{fig:speculative-cases}
\end{figure*}

We examine how information timing and explicit intent revisions affect Heartbeat's speculative tool execution on FDBench v3 as spoken requests evolve.

\paragraph{Information Timing and Intent Revisions}
\label{sec:information_timing}

\begin{table}[!htb]
\centering
\small
\renewcommand{\arraystretch}{1.12}
\begin{tabularx}{\columnwidth}{@{}>{\raggedright\arraybackslash}X>{\centering\arraybackslash}X@{}}
\toprule
Input group & Success rate \\
\midrule
Early information & 73.3\% \\
Late information & 49.2\% \\
Explicit revision & 70.6\% \\
\bottomrule
\end{tabularx}
\caption{Correct tool-call initiation before speech end on FDBench v3, grouped by information timing and explicit intent revision. For revision cases, the rate measures corrected calls following an outdated request. Group definitions are given in Appendix~\ref{app:speculative-details}.}
\label{tab:information_timing}
\end{table}

Across the FDBench groups in Table~\ref{tab:information_timing}, Heartbeat initiates calls consistent with the final request before speech ends in 73.3\% of early-information recordings and 49.2\% when information arrives late. For explicit revisions, 70.6\% contain a corrected call before speech end. Earlier information provides more opportunities for speculative execution, while late constraints and intent revisions can still be incorporated during listening.

\paragraph{Case Studies and Failure Modes}
\label{sec:speculative_cases}

Figure~\ref{fig:speculative-cases} illustrates early order tracking, a late budget constraint, and a search revised from running shoes to hiking boots. Corrected calls do not guarantee consistent use of speculative results: in 4.0\% of recordings, the final answer uses erroneous or outdated results despite a correctly specified request. The cart-update case in Appendix~\ref{app:speculative-details} shows that a corrected search cannot undo an earlier state-changing action. These cases distinguish timely speculative execution from correct use of its results.

\section{Scheduling and Coverage Details}
\label{app:protocol}

\subsection{Overlapping jobs and the response boundary}
\label{app:boundary}
A tool call dispatched during listening contributes its result to the initial Responder context when the result is complete and available at context assembly $c$ (Equation~\ref{eq:availability}). At the detected end of speech, the scheduler stops launching new Thinker invocations and cancels unfinished Thinker and speculative tool tasks, including queued calls. Later results are excluded from the current response.

At context assembly, a completed Thinker record may be available even if its requested tool result is unavailable. Overlapping invocations can finish out of order.

\subsection{Interpreting speculative tool coverage}
\label{app:stc}
Each reference call contributes at most once to a request's coverage fraction. The denominator is the number of reference calls; repeated speculative attempts cannot contribute multiple matches to the same reference call. Matching a reference and detecting a Responder repetition are separate tests. The first establishes relevance under the benchmark's matching rule, while the second removes retained coverage when the Responder makes a call with the same tool name and exactly the same arguments.

Suppose a request requires two calls. Both speculative results are completed and injected, but the Responder repeats one with identical arguments. The request's non-repeated coverage is $1/2$. If one result instead arrives after context assembly and the other is not repeated, the coverage is also $1/2$, but the loss now occurs at the preparation boundary rather than through response-time repetition.

\section{Extended Related Work}\label{app:related-work}

\paragraph{Incremental reasoning and response preparation.}
Preparation from partial input can target intermediate reasoning, candidate responses, or both.
LiveMind stores prefix-based inferences in memory for subsequent reasoning and final response generation \citep{chen2024livemind}.
PredGen instead drafts candidate responses, verifies reusable prefixes as input grows, and prepares audio for the first sentence \citep{li2025predgen}.
LTS-VoiceAgent combines these forms of preparation: its Thinker maintains corrected transcripts, key variables, and plans, while its Speaker uses this state to begin solving and generating responses before speech ends \citep{zou2026ltsvoiceagent}.
The prepared content therefore ranges from reasoning context for later generation to response text and audio that can be reused directly.

\paragraph{Scheduling speculative computation under streaming input.}
Early tool execution was studied in online semantic parsing through prefix-to-program prediction and selective subprogram execution \citep{zhou2022online}.
Subsequent systems differ in how computation proceeds as input changes.
SHANKS interleaves speech chunks with reasoning and tool calls, but its per-chunk generation budget can leave long call arguments incomplete \citep[Appendix~C.2]{chiang2025shanks}.
Stream RAG contrasts fixed-interval parallel retrieval with model-triggered queries that terminate unfinished calls for the previous query \citep{arora2025streamrag}.
In Speculative Interaction Agents (SIA), incoming user or tool updates interrupt ongoing generation, potentially before a complete tool call is produced, while dispatched tools are managed separately \citep{hooper2026sia}.
Outside speech, PASTE predicts upcoming calls from recurring execution patterns and starts them while the LLM is still generating \citep{sui2026paste}.
Heartbeat allows successive Thinker invocations to overlap within a concurrency limit.
This supports denser triggers without making the next input update a deadline for ongoing generation.

\paragraph{Decoupled architectures and coordination interfaces.}
Decoupled systems use different interfaces to pass intermediate work between components.
The \emph{SHANKS + call-after-listen} configuration passes successful calls and their results to a model that processes the complete request \citep{chiang2025shanks}.
DuplexOmni separates real-time interaction from background reasoning and tool use, returning intermediate outputs through a streaming interface \citep{huang2026duplexomni}.
Other interfaces operate within the model's generation process: AsyncLM encodes tool calls and returned results in the token stream \citep{gim2024asynclm}, whereas AsyncReasoning coordinates thinker and writer views through shared model-internal attention state \citep{yakushev2025asyncreasoning}.
Heartbeat connects independent Thinker invocations, asynchronous tool jobs, and an endpoint-triggered Responder through completed reasoning records and tool results.
This external interface permits independent model selection and module-level optimization without requiring shared model-internal state.

\section{Modular Prompt Assembly}
\label{app:prompts}
The following blocks show the shared Thinker and Responder instructions and their message roles. Benchmark-specific instructions, tool schemas, dialogue history, and current-turn state are represented by placeholders.

\newtcblisting{promptcode}{enhanced,breakable,skin first=enhanced,skin middle=enhanced,skin last=enhanced,listing only,overlay={\draw[line width=0.3pt,black] (frame.south west) rectangle (frame.north east);},colback=HBTableGroup,colframe=black,boxrule=0.3pt,sharp corners,left=3pt,right=3pt,top=3pt,bottom=3pt,boxsep=0pt,before skip=6pt,after skip=6pt,listing options={basicstyle=\ttfamily\small,breaklines=true,columns=fullflexible}}

\subsection{Thinker prompt}
\label{app:thinker-prompt}
The Thinker system message declares its role; the user message assembles the fixed preparation rules and variable modules.
\begin{promptcode}
SYSTEM
You are an incremental HeartBeat Thinking worker for a general real-time voice assistant.

USER
[Task]
Work speculatively from the current cumulative partial ASR transcript while the user is speaking. Preserve useful incremental interpretations and propose supported tool calls. You do not answer the user directly.
Response will independently interpret the complete final input.

[Tools]
<Benchmark tool schemas>

[Decision policy]
- Return noop when there is neither a useful new interpretation nor a newly actionable tool call. Repeated speech may still permit a new call when a completed result has supplied a missing dependency.
- Otherwise return think, with one concise incremental operational note and any eligible calls. A new stable intent with missing arguments can produce thinking with an empty tools array. A note that merely says to wait adds no useful interpretation.
- Do not propose a call identical to an existing pending or status-completed HB call. This does not depend on result content or answer sufficiency. A call with different arguments remains eligible when supported by the current request, not merely to bypass this rule.

[Argument rules]
- Copy the exact supplied tool name and schema-defined argument keys. Satisfy required fields, JSON types, enums, and stated preconditions; do not invent tools, fields, or capabilities.
- Support values with the current transcript, unambiguous conversation history, or relevant usable completed results. Do not invent missing facts or fill required fields with placeholders.

[Output format (STRICT)]
Return one JSON object.
{"action":"noop"}
or
{"action":"think","thinking":"...","tools":[{"name":"...","arguments":{...}}]}
Use an empty array when there is no tool call.
No markdown fences, no explanation, no extra keys.

[Extra Instructions]
<HB benchmark instructions, if supplied>

[Conversation History]
<Dialogue history>
[Current Turn State]
<Thinking entries and tool states>
<Optional current-turn metadata>
[Current Fragment]
<Cumulative partial transcript and time range>
\end{promptcode}

\subsection{Responder prompt}
\label{app:responder-prompt}
The Responder system message combines shared continuation rules, tool schemas, and task-specific instructions. Tool schemas are also passed through the structured tool interface.
\begin{promptcode}
SYSTEM
You are continuing work in an existing assistant turn, not starting the user's request over. HeartbeatContext records work already performed for this turn. Your responsibility is to finish the remaining work and give the user the appropriate response.

### Response Core Rules
You are the main response model in a real-time voice assistant. Interpret the complete final user input using the conversation, supplied tool schemas, task instructions, and current-turn HeartbeatContext. The final input takes precedence over an earlier speculative interpretation, including when it corrects or cancels it.

HeartbeatContext is a history, not an action list. Its thinking is speculative: retain useful interpretations, but do not treat intentions as commands, proof of execution, or evidence for missing arguments. An earlier incorrect call does not invalidate another completed call that matches the final request.

Pending, accumulated, and running work is not completion evidence and is not a reason to omit a necessary Response call. Use only result information present in the provided context.

If all required operations are already covered, respond using the available evidence and the task-specific response policy without more calls. Do not expose internal HeartBeat bookkeeping. Do not execute unnecessary operations solely because they appear in earlier thinking.

<Tool schemas>
<Response benchmark instructions and answer format, if supplied>

PREVIOUS MESSAGES
<Dialogue history and tool results>
\end{promptcode}

\subsection{Heartbeat context module}
\label{app:responder-context}
In the illustrated SpeechFCEval configuration, the current user message contains the preparation snapshot, continuation instructions, and complete final transcript. The temporary context is excluded from persistent dialogue history after the response.
\begin{promptcode}
CURRENT USER
### HeartbeatContext
<Actual injected thinking and completed tool records: name, arguments, status, result>

### ResponseConstraint
Continue from the completed work above. Request only new work: never restate a completed exact-name, complete-arguments call in tool_calls. Missing result details do not make that execution pending. If no new operation is needed, produce the task-appropriate response without tool calls. Use the final input to correct genuinely different arguments, not to restart matching completed work.

### FinalUserInput
<Complete final transcript>
\end{promptcode}

\section{Use of AI Assistants}
\label{app:llm-usage}
AI assistants were used for manuscript drafting and revision and for assistance with experimental code. The authors are responsible for the final manuscript, code, analyses, and claims.

\end{document}